\documentclass[letterpaper, 10pt, conference]{ieeeconf}

\IEEEoverridecommandlockouts
\usepackage[utf8]{inputenc}
\usepackage[T1]{fontenc}
\usepackage{amsmath}
\usepackage{amssymb}
\usepackage{graphicx}
\usepackage{booktabs}
\usepackage[hidelinks]{hyperref}
\usepackage{url}
\usepackage{balance}
\usepackage{xcolor}
\usepackage{multirow}
\usepackage{subfig}
\usepackage{bbm}

\title{\LARGE \bf
DAWN: Noise-Robust Quadruped Parkour\\via Depth-Denoising World Models
}

\author{Yohan Choi$^{1}$, Min-Jun Kim$^{1}$, Jin-Sung Kim$^{1}$, Yong-Jae Kim$^{2}$, Youn-Hee Han$^{1,*}$%
\thanks{$^{1}$Future Convergence Engineering, Korea University of Technology and Education, Cheonan 31253, South Korea {\tt\small yoweif@koreatech.ac.kr}, {\tt\small june573166@koreatech.ac.kr}, {\tt\small kjs0820k@koreatech.ac.kr}, {\tt\small yhhan@koreatech.ac.kr}}%
\thanks{$^{2}$School of Electrical, Electronics and Communication Engineering, Korea University of Technology and Education, Cheonan, 31253, South Korea {\tt\small yongjae@koreatech.ac.kr}}%
\thanks{$^{*}$Corresponding Author. {\tt\small yhhan@koreatech.ac.kr}}%
}

\begin{document}

\maketitle
\thispagestyle{empty}
\pagestyle{empty}

\begin{abstract}
Vision-based legged locomotion methods assume clean depth at training time and rely on hand-tuned post-processing filters at deployment.
However, filter parameters are rarely disclosed, hindering reproducibility, and performance degrades substantially when depth noise is left unaddressed.
Building noise robustness directly into the learning pipeline would eliminate this dependency.
While such robustness has been explored for proprioceptive inputs, analogous approaches for depth perception remain largely absent in legged locomotion.
We propose DAWN (Denoising and Alignment in World models for Noise-robustness), a noise-robust perception framework for legged locomotion, which builds noise robustness directly into a world model via two modifications:
(1) feeding noisy depth to the encoder while keeping clean depth as the reconstruction target, forcing the model to implicitly denoise its input; and
(2) applying contrastive learning to align the latent states of noisy and clean depth.
Importantly, DAWN is not tied to a specific noise model, requiring no manual tuning to the noise distribution at deployment. Furthermore, it incurs no additional inference cost over existing world model-based methods.
Without any manual filter calibration—relying solely on the learned noise-robust representation—DAWN achieves zero-shot quadruped parkour on a Unitree Go1: traversing stairs up to 18\,cm, clearing gaps up to 70\,cm, and mounting steps up to 45\,cm from raw depth observations.
Ablation studies show that denoising and contrastive alignment contribute at complementary levels---reconstruction and representation, respectively---and yield additive gains when combined.
Videos and code are available at: \url{https://dawn-parkour.github.io/}
\end{abstract}

\section{INTRODUCTION}
\label{sec:introduction}

Reinforcement learning (RL) has enabled quadruped robots to traverse diverse terrains through sim-to-real transfer~\cite{hwangbo2019agile, tan2018sim2real}.
A blind policy with only proprioceptive input can traverse moderate terrains such as slopes and stairs~\cite{kumar2021rma, lee2020blind}, but fails on obstacles that require perceiving terrain geometry in advance, such as gaps and steps~\cite{agarwal2022ego, miki2022perceptive}.
Depth cameras provide direct geometric information about such terrains and have become the primary sensor for vision-based locomotion~\cite{miki2022perceptive, agarwal2022ego}.
Recent depth-based methods have demonstrated increasingly agile parkour on low-cost quadrupeds, from climbing obstacles up to 0.55\,m and leaping gaps up to 0.85\,m to jumping obstacles exceeding twice the robot's height~\cite{zhuang2023rpl, cheng2024ep, lai2025wmp}.
These and most other depth-based locomotion methods rely on Intel RealSense D435/D435i cameras~\cite{zhuang2023rpl, cheng2024ep, lai2025wmp, hoeller2024anymal, zhuang2024humanoid}. However, the handling of sensor noise at deployment is rarely discussed.

Despite this progress, most vision-based locomotion methods assume clean depth during training and defer noise handling to post-processing filters at deployment~\cite{lai2025wmp, cheng2024ep, zhuang2023rpl, agarwal2022ego}.
Even when depth degradation is acknowledged, common remedies remain outside the main learning pipeline---proprioceptive fallback~\cite{miki2022perceptive} or a separate learned reconstruction module~\cite{hoeller2024anymal}---suggesting that the standard training procedure alone does not produce noise-robust policies.
Moreover, while post-processing filters are widely used at inference~\cite{zhuang2023rpl, cheng2024ep, lai2025wmp, zhuang2024humanoid}, their specific parameters are seldom reported.
The librealsense2 pipeline for the D435i~\cite{keselman2017realsense} exposes six filter types with 12--24 interdependent parameters.
Their optimal values vary with illumination, surface material, and scene depth, making consistent reproduction across environments difficult.

Prior work has quantified the severity of this issue~\cite{sun2026nyst, sun2025dpl}, but the approaches explored so far operate at the input level: domain randomization~\cite{tobin2017dr}, hand-crafted augmentation~\cite{sun2026nyst, sun2025dpl}, or separate denoising modules~\cite{hoeller2024anymal}.
Hand-crafted filters cannot address scene-dependent noise with fixed parameters~\cite{sweeney2019noise}, and prior depth denoising methods exhibit mismatches with current stereo sensors~\cite{hu2023depth}.
These input-level augmentations provide limited representational capacity~\cite{laskin2020rad}.
Meanwhile, world model-based denoising has been applied only to proprioception~\cite{gu2024dwl, sun2025wmr}, where noise is low-dimensional and approximately i.i.d.
Depth images, by contrast, exhibit spatially structured noise whose statistics vary with scene geometry and illumination~\cite{ahn2019d435, keselman2017realsense}---a fundamentally different regime that prior approaches have not addressed.

In this work, we propose DAWN (\textbf{D}enoising and \textbf{A}lignment in \textbf{W}orld models for \textbf{N}oise-robustness), a perception framework for legged locomotion that embeds noise robustness into the latent space of a world model through two modifications to the Recurrent State-Space Model (RSSM) architecture of Lai et al.~\cite{lai2025wmp}.
Specifically, DAWN introduces an input--target mismatch in the RSSM, where the encoder receives noisy depth but the decoder is supervised with clean depth, and applies SimCLR~\cite{chen2020simclr}-based contrastive alignment between the latent representations of noisy and clean depth.
Unlike fixed-parameter filters that require scene-specific tuning, our RSSM encoder learns a nonlinear mapping that discards noise conditioned on scene context, trained end-to-end with the locomotion policy. 
Critically, both modifications apply only at training time, adding no inference cost over the base world model. 
Notably, DAWN does not rely on a specific noise model and needs no tuning to the noise distribution at deployment.

In simulation, DAWN achieves 96.9\% average success rate across stairs,
gaps, and steps from raw depth, closing 77\% of the gap between the
depth-based baseline WMP~\cite{lai2025wmp} and the clean-depth oracle, and degrading
far less than baselines under out-of-distribution noise. Deployed
zero-shot on a real Unitree Go1 without any manual filter calibration,
DAWN traverses stairs up to 18\,cm, gaps up to 70\,cm, and steps up to
45\,cm in both indoor and outdoor environments.

Below, we summarize our main contributions:
\begin{enumerate}
    \item \textbf{DAWN}, a noise-robust perception framework that builds depth denoising directly into the world model, eliminating environment-dependent filter tuning at deployment.

    \item \textbf{Zero-shot sim-to-real quadruped parkour} on a Unitree Go1  across stairs, gaps, and steps, \textbf{without manual filter calibration}.

    \item \textbf{Systematic ablations} confirming that RSSM denoising and contrastive  alignment operate at complementary levels—reconstruction and representation—and  yield additive gains when combined.
\end{enumerate}
These results suggest that noise robustness for depth-based
legged locomotion can be achieved through learned representation rather
than hand-engineered filtering.

\section{RELATED WORK}
\label{sec:related_work}

We review four lines of work relevant to DAWN: visual legged locomotion, depth sim-to-real transfer, world models for locomotion, and contrastive learning for locomotion.

\subsection{Visual Legged Locomotion}
\label{sec:rw_visual}

Sim-to-real RL~\cite{hwangbo2019agile, tan2018sim2real} combined with privileged learning~\cite{chen2019cheating, kumar2021rma} has become the dominant paradigm for legged locomotion.
Policies with only proprioceptive input can traverse moderate terrains such as slopes and stairs~\cite{kumar2021rma}, but fail on obstacles that require perceiving terrain geometry in advance, such as gaps and steps~\cite{agarwal2022ego, miki2022perceptive}.
Depth cameras provide direct geometric information for such terrains, and vision-based parkour has progressed from climbing obstacles up to $1.5\times$ the robot height~\cite{zhuang2023rpl} to jumps exceeding $2\times$ the robot height~\cite{cheng2024ep}.
Lai et al.~\cite{lai2025wmp} introduced a world model-based approach that bypasses the teacher--student distillation pipeline, learning perception and policy end-to-end via the RSSM.
These methods commonly train with clean depth and defer noise handling to hand-tuned post-processing filters at deployment.

\subsection{Depth Sim-to-Real Transfer}
\label{sec:rw_depth}

Domain randomization~\cite{tobin2017dr} is widely used for sim-to-real transfer, but even large-scale automatic domain randomization does not fully close all sim-to-real gaps~\cite{openai2019rubiks}.
Keselman et al.~\cite{keselman2017realsense} reported the official characteristics of the RealSense D400 series, and Ahn et al.~\cite{ahn2019d435} provided an empirical noise model of the D435.
Sweeney et al.~\cite{sweeney2019noise} analyzed scene-dependent filter behavior, and Hu et al.~\cite{hu2023depth} examined the applicability of prior depth denoising methods to current stereo sensors.
Liu et al.~\cite{rasim2024icra} proposed a systematic depth simulation pipeline, and Sun et al.~\cite{sun2025dpl} and Zhuang et al.~\cite{zhuang2024humanoid} introduced physically informed noise synthesis.
Sun et al.~\cite{sun2026nyst} reported a 56\,p.p. success rate gain through eight depth augmentations, quantifying the severity of depth noise.
Hoeller et al.~\cite{hoeller2024anymal} addressed noisy depth with a learned terrain reconstruction module that fuses six depth cameras and LiDAR.
This approach, however, requires auxiliary hardware and is optimized independently of the downstream policy, preventing end-to-end learning.

Existing work focuses on input-level noise modeling or separate denoising modules, both of which require scene-specific filters or auxiliary components at deployment.
Repurposing the reconstruction objective of a world model to remove filter dependency entirely has not been explored.

\subsection{World Models for Locomotion}
\label{sec:rw_world}

Following Ha and Schmidhuber~\cite{ha2018worldmodels}, Hafner et al.~\cite{hafner2019planet} introduced the RSSM, which evolved through subsequent iterations~\cite{hafner2020dreamerv1, hafner2020dreamerv2, hafner2025dreamerv3}.
Wu et al.~\cite{wu2022daydreamer} demonstrated learning locomotion from scratch on a real robot using a world model.

Gu et al.~\cite{gu2024dwl} applied noisy-to-clean reconstruction of proprioception to achieve noise-robust locomotion, and Sun et al.~\cite{sun2025wmr} introduced gradient cutoff to protect denoising quality.
These works, however, address only proprioception, whose noise is low-dimensional and approximately i.i.d.
Depth noise is fundamentally different: it concentrates at depth discontinuities and grows nonlinearly with distance~\cite{ahn2019d435, keselman2017realsense}, requiring architectural considerations for processing spatially structured features.

\subsection{Contrastive Learning for Locomotion}
\label{sec:rw_contrastive}

Contrastive representation learning originated with van den Oord et al.~\cite{oord2018cpc} and advanced in the vision domain through He et al.~\cite{he2020moco} and Chen et al.~\cite{chen2020simclr}.
In RL, Srinivas et al.~\cite{srinivas2020curl} demonstrated improved sample efficiency for pixel-based control.
For locomotion, Long et al.~\cite{long2024him} applied prototypical representation alignment, Mousa et al.~\cite{mousa2025tar} used contrastive triplet loss for teacher--student alignment, and Lu et al.~\cite{lu2025contrastive} improved sim-to-real transfer for humanoid locomotion.
These approaches target proprioceptive or general visual representations; enforcing invariance to depth noise via contrastive learning has not been explored.

\section{METHOD}
\label{sec:method}

\begin{figure*}[!t]
\centering
\includegraphics[width=\textwidth]{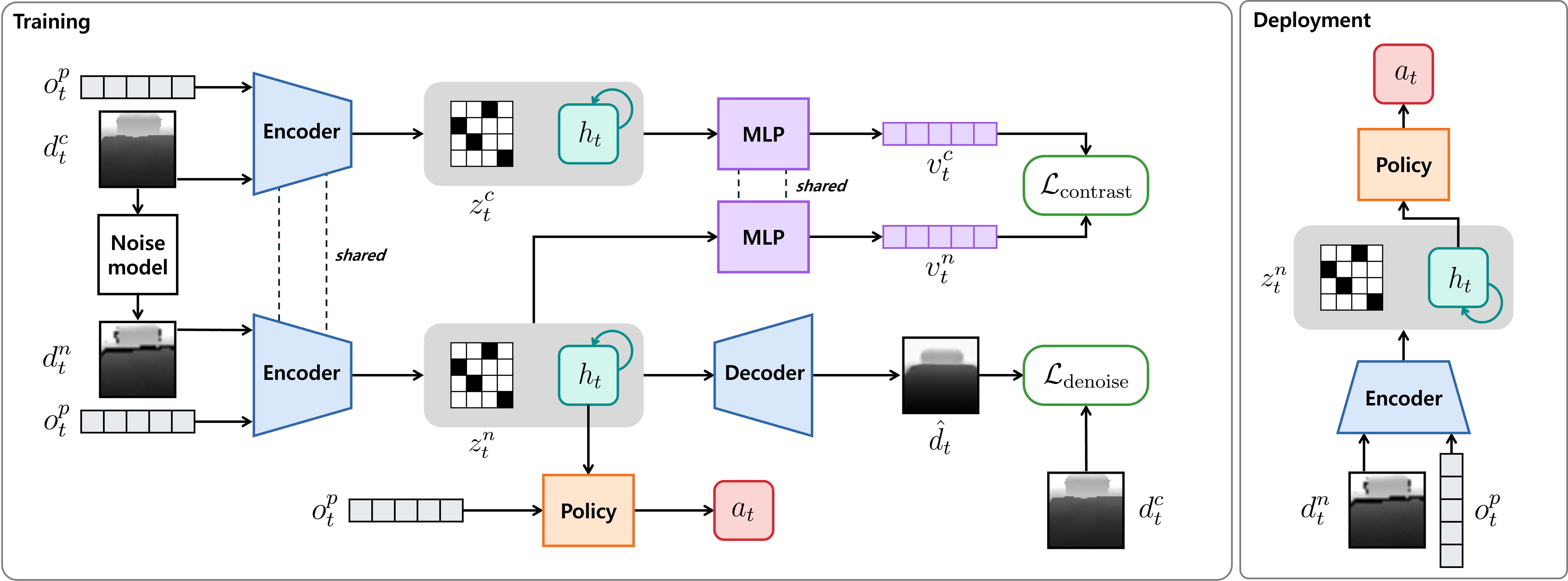}
\caption{Overview of the DAWN framework.
DAWN's two modifications are highlighted in green:
the denoising reconstruction loss $\mathcal{L}_{\text{denoise}}$
between the decoder output and the clean depth target,
and the contrastive loss $\mathcal{L}_{\text{contrast}}$ applied
to the projection head (MLP).}
\label{fig:framework}
\end{figure*}

DAWN builds upon the RSSM architecture of Lai et al.~\cite{lai2025wmp} and introduces two modifications for depth noise robustness:
(1) a denoising reconstruction objective that feeds noisy depth to the encoder while keeping clean depth as the reconstruction target, and
(2) a contrastive loss that aligns latent states from noisy and clean observations.
An overview is shown in Fig.~\ref{fig:framework}.

\subsection{Preliminaries: World Model Learning}
\label{sec:method_preliminaries}

We define the observation at timestep $t$ as $x_t = (d_t, o^{\text{p}}_t)$, where $d_t$ is the depth image and $o^{\text{p}}_t$ is the proprioception.
The depth image is updated every $k$ steps, and $a_t$ denotes the joint position target action.

Lai et al.~\cite{lai2025wmp} propose World Model-based Perception~(WMP), an end-to-end framework that adopts an RSSM following Hafner et al.~\cite{hafner2020dreamerv2}.
The RSSM consists of four components parameterized by $\phi$:
\begin{align}
\text{Sequence model:} \quad h_t &= f_\phi(h_{t-k},\, z_{t-k},\, a_{t-k:t-1}) \label{eq:rssm_seq} \\
\text{Encoder:} \quad z_t &\sim q_\phi(\cdot \mid h_t,\, x_t) \label{eq:rssm_enc} \\
\text{Dynamics predictor:} \quad \hat{z}_t &\sim p_\phi(\cdot \mid h_t) \label{eq:rssm_prior} \\
\text{Decoder:} \quad \hat{x}_t &\sim p_\phi(\cdot \mid h_t,\, z_t) \label{eq:rssm_dec}
\end{align}
where $h_t$ is a deterministic state computed by a Gated Recurrent Unit (GRU)~\cite{cho2014gru}-based sequence model~\eqref{eq:rssm_seq}, $z_t$ is the stochastic state incorporating the current observation~\eqref{eq:rssm_enc}, $\hat{z}_t$ is the prior predicted without observation access~\eqref{eq:rssm_prior}, and $\hat{x}_t$ is the reconstructed observation~\eqref{eq:rssm_dec}.

These components are jointly optimized by minimizing:
\begin{align}
\mathcal{L}_{\text{WMP}}
= \mathbb{E}\!\Bigl[\sum_t \;
  &\underbrace{-\ln p_\phi(x_t \mid z_t,\, h_t)}_{\text{reconstruction}} \notag \\
  + \; &\beta\,\mathrm{KL}\!\bigl[q_\phi(z_t \mid h_t,\, x_t)
        \,\|\, p_\phi(z_t \mid h_t)\bigr]
\Bigr] ,
\label{eq:loss_wmp}
\end{align}
where $\beta$ is a hyperparameter.
The reconstruction term encourages $z_t$ to retain sufficient information about $x_t$, while the KL term regularizes the posterior toward the prior.
The policy is trained with Proximal Policy Optimization~\cite{schulman2017ppo} using the deterministic state $h_t$ as input.

In WMP, both the encoder input and the reconstruction target are the \emph{same} clean observation $x_t$.
Depth noise at deployment is handled by post-processing filters applied to raw sensor readings.
DAWN removes this filter dependency by modifying how the RSSM is trained, without changing its architecture.

\subsection{Denoising Reconstruction Objective}
\label{sec:method_denoising}

We denote clean and noisy depth as $d^{\text{c}}_t$ and $d^{\text{n}}_t$, respectively, and write the corresponding observations as
$x^{\text{c}}_t = (d^{\text{c}}_t,\, o^{\text{p}}_t)$ and $x^{\text{n}}_t = (d^{\text{n}}_t,\, o^{\text{p}}_t)$.
The noisy depth $d^{\text{n}}_t$ is generated by a noise model.

In WMP's standard training, the encoder input and the decoder target are identical:
\begin{equation}
z_t \sim q_\phi(\cdot \mid h_t,\, x^{\text{c}}_t),
\qquad
\hat{x}_t \approx x^{\text{c}}_t .
\label{eq:wmp_io}
\end{equation}
DAWN replaces the encoder input with the noisy observation while keeping the clean observation as the reconstruction target:
\begin{equation}
z^{\text{n}}_t \sim q_\phi(\cdot \mid h_t,\, x^{\text{n}}_t),
\qquad
\hat{x}_t \approx x^{\text{c}}_t .
\label{eq:dawn_io}
\end{equation}
Comparing~\eqref{eq:wmp_io} and~\eqref{eq:dawn_io}, the only change is in the encoder input: from $x^{\text{c}}_t$ to $x^{\text{n}}_t$.
The decoder must still reconstruct the clean observation, which forces the encoder to map noisy input to a latent state from which clean depth can be recovered.

This input--target mismatch turns the standard reconstruction loss into a denoising objective.
Substituting~\eqref{eq:dawn_io} into the WMP loss~\eqref{eq:loss_wmp} yields the DAWN reconstruction loss:
\begin{align}
\mathcal{L}_{\text{denoise}}
= \mathbb{E}\!\Bigl[\sum_t \;
  &\underbrace{-\ln p_\phi(x^{\text{c}}_t \mid z^{\text{n}}_t,\, h_t)}_{\text{denoising}} \notag \\
  + \; &\beta\,\mathrm{KL}\!\bigl[q_\phi(z^{\text{n}}_t \mid h_t,\, x^{\text{n}}_t)
        \,\|\, p_\phi(z_t \mid h_t)\bigr]
\Bigr] .
\label{eq:loss_denoise}
\end{align}
The KL term constrains the information capacity of $z^{\text{n}}_t$.
Since reconstructing $x^{\text{c}}_t$ does not require any information about the noise $\epsilon$ in $x^{\text{n}}_t$, the encoder discards noise-specific features under this capacity constraint and retains only terrain-relevant geometry.
This behavior follows directly from the information bottleneck principle~\cite{tishby2000ib}: when the target is $x^{\text{c}}_t$, the mutual information $I(z^{\text{n}}_t;\, \epsilon)$ does not contribute to reducing the reconstruction loss and is suppressed by the KL penalty.

This learned compression provides a practical advantage over hand-crafted post-processing filters.
Filter pipelines operate with a fixed set of parameters that require scene-specific tuning, whereas the encoder learns a nonlinear mapping trained end-to-end with the policy and conditioned on scene context.
Because the denoising objective already drives the encoder to retain only task-relevant geometry, no additional filter stage is required at deployment.

\begin{table}[t]
\centering
\caption{Comparison of WMP and DAWN training configurations. The RSSM architecture and policy training are identical; only the encoder input and loss terms differ.}
\label{tab:comparison}
\begin{tabular}{lcc}
\toprule
 & WMP & DAWN \\
\midrule
Encoder input & $x^{\text{c}}_t$ & $x^{\text{n}}_t$ \\
Decoder target & $x^{\text{c}}_t$ & $x^{\text{c}}_t$ \\
Reconstruction loss & $\mathcal{L}_{\text{WMP}}$ & $\mathcal{L}_{\text{denoise}}$ \\
Contrastive loss & -- & $\mathcal{L}_{\text{contrast}}$ \\
Post-processing filter & Required & Not required \\
\bottomrule
\end{tabular}
\end{table}

\subsection{Contrastive Latent Alignment}
\label{sec:method_contrastive}

The denoising objective encourages noise-invariant reconstruction but does not explicitly constrain the latent space structure.
We add a contrastive loss to directly align the encoder outputs from clean and noisy observations.

We write $z^{\text{c}}_t \sim q_\phi(\cdot \mid h_t, x^{\text{c}}_t)$ and $z^{\text{n}}_t \sim q_\phi(\cdot \mid h_t, x^{\text{n}}_t)$ for the posterior states encoded from clean and noisy inputs at the same timestep.
A projection head $g_\psi$ maps these posteriors to a contrastive embedding space:
$v^{\text{c}}_t = g_\psi(h_t,\, z^{\text{c}}_t)$ and $v^{\text{n}}_t = g_\psi(h_t,\, z^{\text{n}}_t)$.

From $N$ parallel environments within a batch,
the clean and noisy embeddings of the same scene $i$,
$(v_i^{\text{c}},\, v_i^{\text{n}})$, form a positive pair,
while those from different scenes $i \neq j$ form negative pairs.
The Normalized Temperature-scaled Cross Entropy (NT-Xent) loss is:
\begin{equation}
\mathcal{L}_{\text{contrast}} = -\log \frac{\exp(\mathrm{sim}(v^{\text{c}}_i,\, v^{\text{n}}_i) / \tau)}
{\displaystyle\sum_{j=1}^{2N} \mathbf{1}_{[j \neq i]}\,
\exp(\mathrm{sim}(v^{\text{c}}_i,\, v_j) / \tau)} ,
\label{eq:loss_contrast}
\end{equation}
where $v^{\text{c}}_i$ is the anchor,
$\mathrm{sim}(\cdot,\cdot)$ denotes cosine similarity,
$\mathbf{1}_{[j \neq i]}$ is an indicator function excluding the anchor,
$\{v_j\}_{j=1}^{2N} = \{v^{\text{c}}_1, \dots, v^{\text{c}}_N,\, v^{\text{n}}_1, \dots, v^{\text{n}}_N\}$
is the set of all embeddings in the batch,
and $\tau$ is the temperature parameter.

Following Chen et al.~\cite{chen2020simclr}, the contrastive loss is applied to the projected embeddings $v_t$ rather than directly to $z_t$, preventing the encoder from collapsing toward the contrastive objective and preserving diverse information useful for policy learning.
At deployment, the projection head $g_\psi$ is removed.
The contrastive loss serves only as a training-time regularizer that shapes the latent geometry; the inference pipeline remains identical to WMP.

\subsection{Total Training Objective}
\label{sec:method_total}

The full DAWN loss combines the denoising reconstruction objective~\eqref{eq:loss_denoise} with the contrastive alignment loss~\eqref{eq:loss_contrast}:
\begin{equation}
\mathcal{L}_{\text{DAWN}}
= \mathcal{L}_{\text{denoise}} + \lambda\,\mathcal{L}_{\text{contrast}} ,
\label{eq:loss_total}
\end{equation}
where $\lambda$ controls the relative weight of the contrastive term.
Setting $\lambda = 0$ and replacing $x^{\text{n}}_t$ with $x^{\text{c}}_t$ in~\eqref{eq:loss_denoise} recovers the original WMP loss~\eqref{eq:loss_wmp}.

Table~\ref{tab:comparison} summarizes the differences between WMP and DAWN.
The RSSM architecture and the policy training procedure remain unchanged; DAWN modifies only the training signals.

\begin{table*}[!t]
\centering
\caption{Baseline comparison (noise $\times 1.0$). SR (\%, $\uparrow$) and TE (m/s, $\downarrow$) are reported.
Averaged over all difficulty levels, 3 seeds $\times$ 100 episodes.}
\label{tab:baseline}
\normalsize
\resizebox{\textwidth}{!}{%
\begin{tabular}{l cc cc cc cc}
\toprule
& \multicolumn{2}{c}{Slope (6--22$^\circ$)}
& \multicolumn{2}{c}{Stair (6--18\,cm)}
& \multicolumn{2}{c}{Gap (10--90\,cm)}
& \multicolumn{2}{c}{Step (10--54\,cm)} \\
\cmidrule(lr){2-3} \cmidrule(lr){4-5} \cmidrule(lr){6-7} \cmidrule(lr){8-9}
Method & SR & TE & SR & TE & SR & TE & SR & TE \\
\midrule
Blind         & $98.9{\pm 3.8}$          & $0.025{\pm .004}$          & $74.6{\pm 44.5}$          & $0.064{\pm .066}$          & $26.2{\pm 44.0}$          & $0.154{\pm .073}$ & $36.2{\pm 43.4}$          & $0.936{\pm .071}$ \\
EP            & $100.0{\pm 0.0}$         & $0.011{\pm .003}$          & $82.2{\pm 35.4}$          & $0.031{\pm .042}$          & $83.5{\pm 35.9}$          & $0.072{\pm .048}$ & $92.6{\pm 23.2}$          & $0.038{\pm .031}$ \\
WMP           & $100.0{\pm 0.0}$         & $0.010{\pm .003}$          & $89.7{\pm 30.4}$          & $0.026{\pm .038}$          & $88.6{\pm 31.8}$          & $\mathbf{0.058}{\pm .034}$ & $95.9{\pm 19.8}$          & $0.026{\pm .018}$ \\
WMP w/ N      & $99.9{\pm 3.2}$          & $0.009{\pm .003}$          & $94.5{\pm 22.8}$          & $0.023{\pm .036}$          & $93.7{\pm 24.3}$          & $0.067{\pm .029}$ & $95.3{\pm 21.2}$          & $\mathbf{0.025}{\pm .021}$ \\
\textbf{DAWN} & $\mathbf{99.9}{\pm 2.5}$ & $\mathbf{0.008}{\pm .002}$ & $\mathbf{96.6}{\pm 18.2}$ & $\mathbf{0.015}{\pm .020}$ & $\mathbf{97.2}{\pm 16.6}$ & $0.065{\pm .031}$ & $\mathbf{97.0}{\pm 17.0}$ & $0.034{\pm .020}$ \\
Oracle        & $100.0{\pm 0.0}$         & $0.008{\pm .002}$          & $98.2{\pm 13.3}$          & $0.013{\pm .015}$          & $98.8{\pm 10.9}$          & $0.054{\pm .028}$ & $98.5{\pm 12.2}$          & $0.020{\pm .014}$ \\
\bottomrule
\end{tabular}}
\end{table*}

\begin{table}[!b]
\centering
\caption{Summary of compared methods. Noise: depth noise applied during training. Denoise: reconstruction target set to clean depth. Contrastive: contrastive learning applied.}
\label{tab:methods}
\normalsize
\begin{tabular}{lccc}
\toprule
Method & Noise & Denoise & Contrastive \\
\midrule
Blind & -- & -- & -- \\
EP~\cite{cheng2024ep} & \texttimes & \texttimes & \texttimes \\
WMP~\cite{lai2025wmp} & \texttimes & \texttimes & \texttimes \\
WMP w/ N & \checkmark & \texttimes & \texttimes \\
DAWN w/o C & \checkmark & \checkmark & \texttimes \\
DAWN w/o D & \checkmark & \texttimes & \checkmark \\
\textbf{DAWN} & \checkmark & \checkmark & \checkmark \\
Oracle & -- & -- & -- \\
\bottomrule
\end{tabular}
\end{table}

\section{EXPERIMENTAL RESULTS}
\label{sec:experiments}

\subsection{Experimental Setup}
\label{sec:exp_setup}

\textbf{Simulation.}
Training is conducted in IsaacLab~\cite{mittal2025isaaclab} with 4,096 parallel environments.
The control frequency is 50\,Hz, and depth images at $64 \times 64$ resolution are updated every $k = 5$ steps (0.1\,s).
All experiments are repeated with three seeds, and 100 episodes per condition are evaluated, reporting mean$\pm$standard deviation.

\textbf{Real-world.}
A Unitree Go1 is equipped with an Intel RealSense D435i depth camera and an Nvidia Jetson NX, with all RealSense built-in filters disabled so that the policy receives raw depth.
Policies trained in simulation are transferred zero-shot without additional fine-tuning.
We evaluate the success rate over 10 trials per difficulty level across both indoor and outdoor environments.

\textbf{Depth Noise Simulation.}
DAWN does not require tuning to a specific noise model.
We use a D435i-characteristic model~\cite{ahn2019d435} capturing three
phenomena, validated in \S\ref{sec:exp_noise_valid}.
We denote the intermediate depth image as $\tilde{d}$.
The noise model applies the following filters to $d^{\text{c}}_t$ to
produce the final noisy depth $d^{\text{n}}_t$:

\begin{itemize}
\item \textit{Gaussian sensor noise} adds zero-mean noise:
\begin{equation}
  \tilde{d} = d^{\text{c}}_t + \eta, \quad
  \eta \sim \mathcal{N}(0,\,\sigma^2), \quad
  \sigma = 0.01\text{\,m}.
  \label{eq:gaussian_noise}
\end{equation}

\item \textit{Edge dropout} simulates stereo matching failure at
depth discontinuities with quadratic probability:
\begin{equation}
  P_{\text{drop}}(x,y) = P_{\max} \cdot
  \text{clip}\!\left(
    \frac{G - G_{\text{th}}}{G_{\text{sat}} - G_{\text{th}}},\,0,\,1
  \right)^{\!2},
  \label{eq:edge_dropout}
\end{equation}
where $G = \|\nabla \tilde{d}\|$ is the spatial gradient magnitude,
$G_{\text{th}} = 0.04$ is the minimum gradient for edge detection,
$G_{\text{sat}} = 0.20$ is the gradient at which the dropout probability
saturates, and $P_{\max} = 0.60$ is the upper bound on dropout probability.
Dropped pixels are filled by $3 \times 3$ max pooling.

\item \textit{Far particle noise} models infrared
(IR)--ambient light interference:
\begin{equation}
  P_{\text{particle}}(x,y) = r_p \cdot
  \text{clip}\!\left(
    \frac{\tilde{d} - d_{\text{th}}}{0.2},\,0,\,1
  \right),
  \label{eq:far_particle}
\end{equation}
where $r_p = 0.003$ is the maximum particle rate and
$d_{\text{th}} = 0.3$ is the depth threshold beyond which
particles are applied.
\end{itemize}

\textbf{Terrains.}
We use four terrain types, each with five difficulty levels: Slope (6--22$^\circ$), Stair (6--18\,cm), Gap (10--90\,cm), and Step (10--54\,cm).
Slope is included only in the baseline comparison (Table~\ref{tab:baseline}) and excluded from ablation and noise sweep analyses, as its lack of depth discontinuities yields similar performance across all methods.

\textbf{Compared Methods.}
Table~\ref{tab:methods} summarizes the compared methods.
WMP w/ N adds noise training to WMP. DAWN w/o C and DAWN w/o D are ablation variants that remove contrastive learning and RSSM denoising, respectively, yielding denoising-only and contrastive-only variants.

\textbf{Metrics.}
We report Success Rate (SR, \%) and velocity Tracking Error (TE, m/s).
SR measures the percentage of episodes in which the robot successfully
traverses the terrain without falling.
TE is the mean squared error between the commanded velocity and the measured velocity capturing how precisely the robot tracks the desired motion.

\subsection{Baseline Comparison}
\label{sec:exp_baseline}

Table~\ref{tab:baseline} reports baseline results under the default $\times 1.0$ noise, averaged across all difficulty levels over three seeds.
We use this comparison to answer whether depth denoising must be built into the learning pipeline or can be deferred to the training data and to post-processing filters.

The terrain-wise pattern isolates where noise hurts.
Slope yields near-perfect SR for every method, including Blind: with no depth discontinuities, proprioception alone suffices and depth noise is inconsequential.
The picture reverses on terrains defined by geometric edges.
Blind collapses on Gap and Step, confirming that advance depth perception is required once the robot must anticipate an obstacle it cannot feel.
WMP, the clean-depth world-model baseline we build on, degrades most on Stair and Gap, where edge dropout removes depth precisely at the boundaries the policy relies on—Stair through consecutive edges subject to cumulative dropout, Gap through a width estimate that needs both rims visible at once.
Step is more forgiving, since a single discontinuity leaves the height change recoverable from depth values on either side.
EP, the distillation-based policy, trails WMP on every perceptive terrain, indicating that the teacher–student pipeline carries less usable geometry to the policy than the end-to-end world model even before noise is considered.

Across Stair/Gap/Step, DAWN reaches 96.9\% average SR, closing 77\% of the gap that noise opens between WMP and the clean-depth Oracle.
WMP w/ N recovers part of this gap through noise exposure alone but stays below DAWN, showing that seeing noise during training is not equivalent to structuring the representation against it.
The trend carries to tracking: DAWN attains the lowest TE on Slope and Stair, cutting WMP's Stair error by 42\%, which we attribute to denoised depth producing more precise foot placement.
On Gap and Step, TE differences among the world-model methods fall within overlapping variance, indicating that under ×1.0 noise the corrupted boundaries surface primarily in SR rather than in tracking.

\begin{figure*}[!t]
\centering
\includegraphics[width=0.33\textwidth]{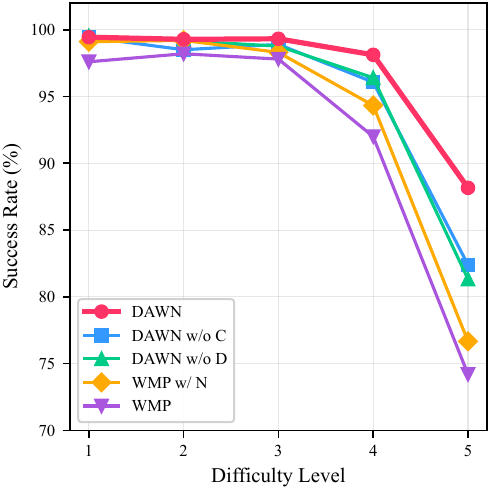}%
\hfill
\includegraphics[width=0.33\textwidth]{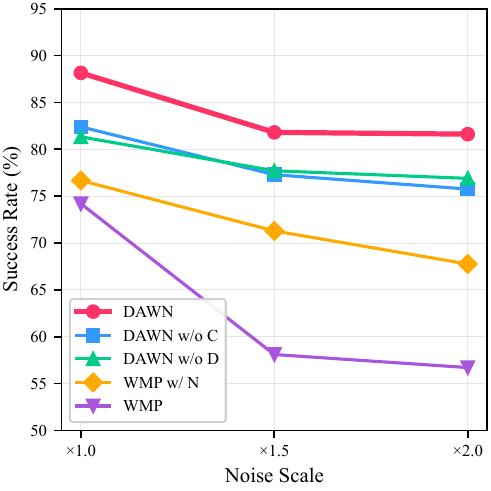}%
\hfill
\includegraphics[width=0.33\textwidth]{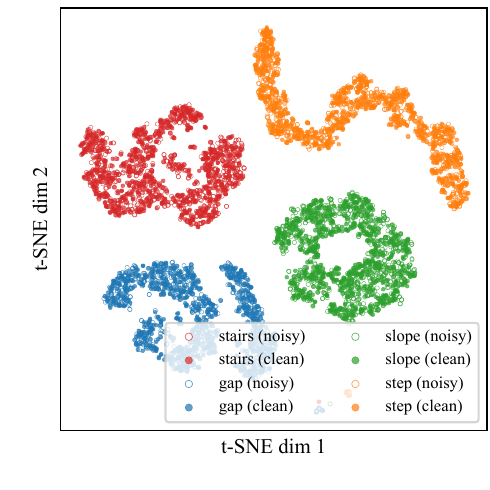}
\caption{Simulation analysis.
Ablation study showing average SR over Stairs/Gap/Step by difficulty level (noise ×1.0), mean of 3 seeds (left). Noise robustness analysis showing average SR over Stairs/Gap/Step at the
highest difficulty as the noise scale increases from $\times 1.0$ to
$\times 2.0$, averaged over 3 seeds (center). t-SNE visualization of encoder latent states; colors indicate terrain type, shading indicates input condition (filled: clean, open: noisy) (right).
}
\label{fig:analysis}
\end{figure*}

\begin{figure}[!t]
\centering
\includegraphics[width=\columnwidth]{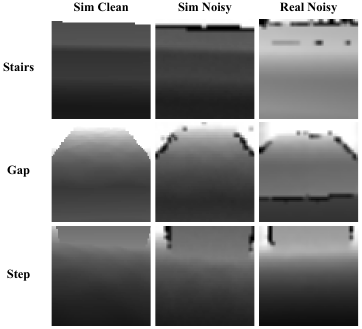}
\caption{Comparison of simulated clean depth, simulated noisy depth ($\times 1.0$), and real D435i depth across three terrains (Stairs, Gap,  Step).
The simulated noise patterns closely match real sensor output.}
\label{fig:sim_real}
\end{figure}

\begin{figure}[!t]
\centering
\includegraphics[width=\columnwidth]{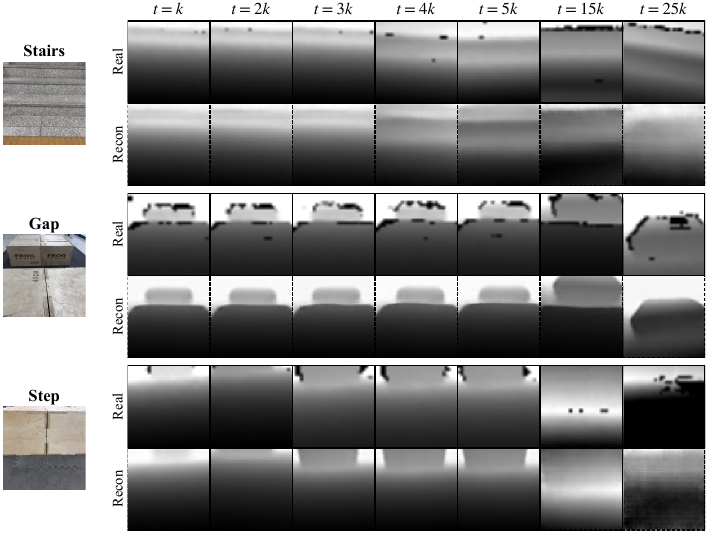}
\caption{Real-world depth reconstruction on three terrains (Stairs, Gap, Step). For each terrain, the top row is the raw D435i depth input and the bottom row (dashed border) is DAWN's decoder output at the same timestep, shown at t = k, 2k, …, 5k, 15k, and 25k.}
\label{fig:reconstruction}
\end{figure}

\begin{table}[!t]
\centering
\caption{Noise scale parameters. $\times 1.0$ is the default setting.}
\label{tab:noise_scale}
\normalsize
\begin{tabular*}{\columnwidth}{l@{\extracolsep{\fill}}ccccc}
\toprule
Parameter & $\times 1.0$ & $\times 1.5$ & $\times 2.0$ \\
\midrule
Gaussian $\sigma$ (m) & \textbf{0.01} & 0.015 & 0.02 \\
Edge $P_{\max}$ & \textbf{0.60} & 0.80 & 0.90 \\
Particle $r_p$ & \textbf{0.003} & 0.0045 & 0.006 \\
\bottomrule
\end{tabular*}
\end{table}

\begin{figure*}[!t]
\centering
\includegraphics[width=\textwidth]{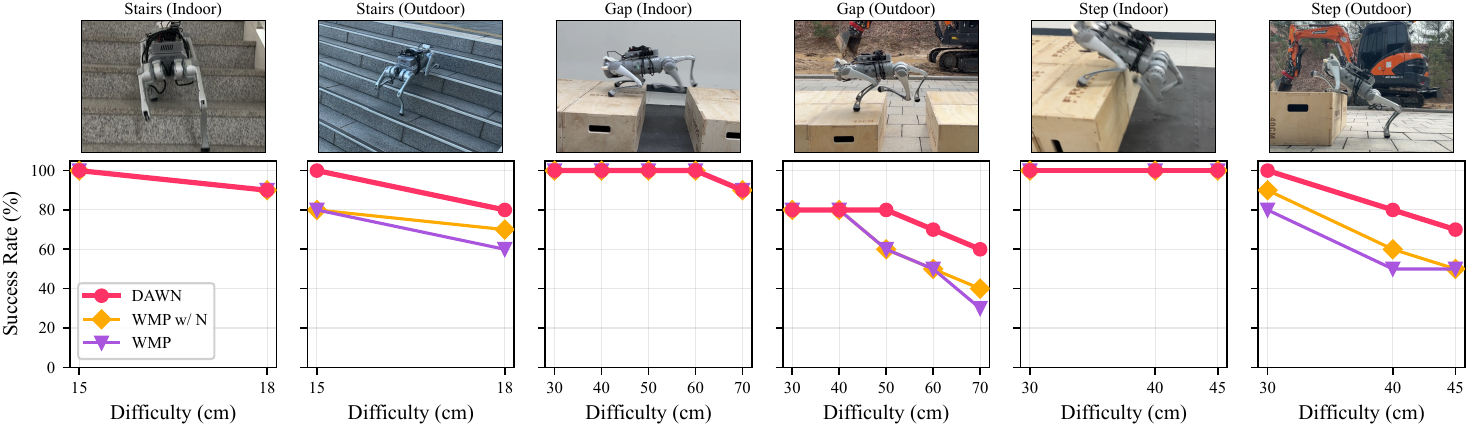}
\caption{Real-world experimental results.
Three methods (DAWN, WMP w/ N, WMP) $\times$ three terrains (Stairs, Gap, Step) $\times$ two environments (Indoor, Outdoor).
SR (\%) over 10 trials per difficulty level.}
\label{fig:realworld}
\end{figure*}

\subsection{Ablation Study}
\label{sec:exp_ablation}

Fig.~\ref{fig:analysis} (left) isolates the contribution of each modification by tracking average SR over Stair/Gap/Step as difficulty increases.
We compare five methods: the full DAWN; its two single-component variants (DAWN w/o C, denoising only; DAWN w/o D, contrastive only); and two WMP baselines that differ only in noise training—WMP w/ N, trained with noisy depth, and WMP, trained on clean depth.

All five methods perform near-ceiling through the low difficulties, where obstacles are small and the surviving depth is sufficient regardless of how it is processed.
The methods separate only from level 4 onward: larger obstacles subtend wider depth discontinuities, producing broader dropout regions that make the quality of the recovered geometry the limiting factor.
We therefore read the ablation at the hardest level, where the noise stress is greatest.
There, DAWN reaches 88.2\% SR.
Measured as increments over noise-only training, denoising alone adds 5.7 p.p. and contrastive alignment alone 4.6 p.p., while the two together add 11.5 p.p.—more than the sum of the parts, so the components are complementary rather than redundant.

The complementarity follows from how the two objectives interact: denoising restructures the latent space toward noise-free geometry, giving contrastive alignment a stable target to match, while alignment in turn regularizes the encoder output so the decoder reconstructs from a cleaner state.
The two objectives provide additive performance gains when combined.

\subsection{Noise Robustness Analysis}
\label{sec:exp_noise}

The $\times 1.0$ results establish robustness at the noise level seen during training.
Fig.~\ref{fig:analysis} (center) sweeps the noise scale from $\times$1.0 to $\times$2.0 at the hardest difficulty (see Table~\ref{tab:noise_scale}); since all policies are trained at $\times$1.0, the higher scales are out-of-distribution.
Across the sweep DAWN loses 6.5\,p.p.\ versus 17.5 for clean-trained WMP, and the DAWN--WMP margin widens from 14.0\,p.p.\ at $\times$1.0 to 24.9\,p.p.\ at $\times$2.0, with the DAWN--WMP~w/~N margin growing in step: the baselines
lose ground fastest exactly where noise is most severe, whereas DAWN's
advantage grows.
We attribute this to where each method acts: domain randomization fits the encoder to the noise levels it has seen and extrapolates poorly beyond them, while denoising and contrastive alignment reshape the latent space toward noise-free geometry—a target that does not move as the input noise intensifies.
This margin behavior anticipates the real-world results (\S\ref{sec:exp_realworld}), where outdoor IR interference pushes sensor noise past the indoor regime and the clean-trained baselines degrade most.

\subsection{Latent Space Analysis}
\label{sec:exp_latent}

Fig.~\ref{fig:analysis} (right) shows a t-SNE visualization of the concatenated $(h_t, z_t)$.
For each terrain type, the clean and noisy clusters overlap, indicating that the encoder maps both to overlapping regions of the latent space rather than separating them by noise condition.
This means the encoder has learned to discard depth noise and extract only terrain geometry, achieving the noise invariance that DAWN targets.
Meanwhile, different terrain types form well-separated clusters, confirming that the encoder preserves terrain discriminability.

\subsection{Noise Model Validation}
\label{sec:exp_noise_valid}

Fig.~\ref{fig:sim_real} compares simulated clean depth, simulated noisy depth ($\times 1.0$), and real D435i depth across three terrain views.
The simulated noise visually matches real sensor output, particularly the missing pixels at depth discontinuities and far particle artifacts at distance, validating that the noise model captures the dominant characteristics of the D435i.

\subsection{Real-World Depth Reconstruction}
\label{sec:exp_reconstruction}

Fig.~\ref{fig:reconstruction} shows DAWN's RSSM decoder reconstructing clean depth from real D435i input across consecutive timesteps on Stair, Gap, and Step.
This confirms that the deterministic state $h_t$ has learned to discard sensor noise.
Since the policy conditions on $h_t$, it operates on noisy observations without requiring external filters.

\subsection{Real-World Deployment}
\label{sec:exp_realworld}

\begin{figure}[t]
\centering
\includegraphics[width=\columnwidth]{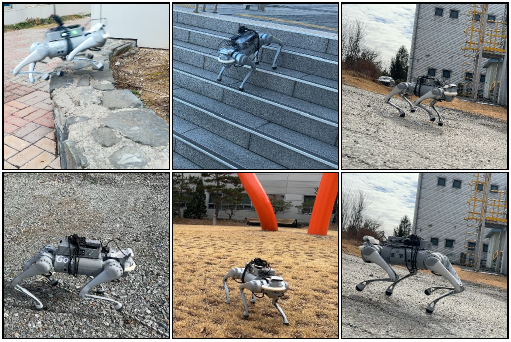}
\caption{Outdoor deployment snapshots: curb climbing on stone terrain,
stair descending, slope descending, gravel walking, grass walking,
and slope ascending.}
\label{fig:outdoor}
\end{figure}

Fig.~\ref{fig:realworld} presents real-robot results on a Unitree Go1
in indoor and outdoor environments, with SR measured over 10 trials
per difficulty level. In indoor environments, all three methods perform comparably, holding near 100\% SR and dropping to 90\% only at the hardest Stair (18 cm) and Gap (70 cm); Step shows no degradation at any level.
The performance gap emerges outdoors, where sunlight-induced IR
interference and surface material variation amplify depth corruption.
On Stair at 18\,cm, DAWN achieves 80\% vs.\ WMP w/ N 70\% and WMP 60\%.
On Gap at 70\,cm, DAWN reaches 60\%, WMP w/ N 40\%, and WMP 30\%.
On Step at 45\,cm, DAWN records 70\% vs.\ WMP w/ N and WMP both at 50\%.
This preserves the three-way ordering observed in simulation (\S\ref{sec:exp_noise}): DAWN leads, noise-trained WMP w/ N sits between, and clean-trained WMP trails.
Noise exposure during training therefore helps outdoors but does not substitute for DAWN's noise-robust representation, which absorbs the added IR and surface-material corruption without any filter tuning.
Fig.~\ref{fig:outdoor} shows qualitative snapshots of outdoor deployment across diverse
terrains and surface materials.

\section{CONCLUSION}
\label{sec:conclusion}

We presented DAWN, a noise-robust perception framework for quadruped parkour that builds depth noise robustness into the world model's existing mechanisms, requiring no external modules or environment-specific tuning.
DAWN passes noisy depth to the RSSM encoder while keeping clean depth
as the reconstruction target, forcing the model to implicitly denoise
its input.
A contrastive loss further aligns noisy and clean representations at
the encoder level.
Together, these modifications remove the need for environment-dependent
filter tuning at deployment while adding no inference overhead.
In simulation, DAWN achieved 96.9\% average success rate across stairs, gaps, and steps---close to the clean-depth Oracle---and these improvements transferred consistently to a real Unitree Go1 via zero-shot deployment.
Ablation confirmed that the two modifications serve complementary roles: denoising restructures the representation space toward clean geometry, while contrastive learning enforces noise-robust encoding, yielding a compound gain when combined.
The current noise model targets the D435i; extending it to other depth sensors is a natural direction for future work.

\section*{Acknowledgment}
\label{sec:acknowledgment}

This research was supported by Basic Science Research Program through the National Research Foundation of Korea (NRF) funded by the Ministry of Education (2018R1A6A1A03025526).

\balance
\bibliographystyle{IEEEtran}
{\footnotesize \bibliography{references}}

\end{document}